\documentclass[twoside,leqno,twocolumn]{article}

\usepackage[letterpaper]{geometry}

\usepackage{ltexpprt}
\usepackage{hyperref}

\usepackage{subfig}
\usepackage{amsmath, amssymb}
\usepackage{amsfonts}
\usepackage{physics}
\usepackage{bm}
\usepackage{mathtools}
\usepackage{enumitem}
\mathtoolsset{showonlyrefs}
\usepackage{algorithm}
\usepackage{algpseudocode}
\usepackage{comment}

\newcommand{\Transpose}{^{\mathsf{T}}}

\newcommand{\HiddenStateSymb}{h}
\newcommand{\HiddenState}{\bm{\HiddenStateSymb}}

\newcommand{\NonlinearActivation}{\sigma}

\newcommand{\LatentStateSymb}{z}
\newcommand{\LatentState}{\bm{\LatentStateSymb}}

\newcommand{\NumLatentState}{n_{\LatentStateSymb}}

\newcommand{\nTrain}{n_{\text{train}}}

\newcommand{\VelocitySymb}{f}
\newcommand{\Velocity}{\bm{\VelocitySymb}}

\newcommand{\FlowMapSymb}{F}
\newcommand{\FlowMap}{\bm{\FlowMapSymb}}

\newcommand{\NNParamsSymb}{\Theta}
\newcommand{\NNParams}{\bm{\NNParamsSymb}}

\newcommand{\NNWeightSymb}{W}
\newcommand{\NNWeight}{\NNWeightSymb}

\newcommand{\NNBiasSymb}{b}
\newcommand{\NNBias}{\bm{\NNBiasSymb}}

\newcommand{\TimeSeriesSymb}{x}
\newcommand{\TimeSeries}{\bm{\TimeSeriesSymb}}
\newcommand{\TimeSeriesSevenOTwo}{\TimeSeries^{(702)}}
\newcommand{\TimeSeriesNineOOne}{\TimeSeries^{(901)}}
\newcommand{\TimeSeriesNineOneOne}{\TimeSeries^{(911)}}

\newcommand{\DotTimeSeriesSevenOTwo}{\dot{\TimeSeries}^{(702)}}
\newcommand{\DotTimeSeriesNineOOne}{\dot{\TimeSeries}^{(901)}}
\newcommand{\DotTimeSeriesNineOneOne}{\dot{\TimeSeries}^{(911)}}

\newcommand{\prunethreshold}{\rho}

\usepackage{balance}

\begin{document}

\title{Mining Causality from Continuous-time Dynamics Models: \\An Application to Tsunami Forecasting}

\author{Fan Wu\thanks{Arizona State University,  fanwu8@asu.edu}
\and Sanghyun Hong\thanks{Oregon State University, sanghyun.hong@oregonstate.edu}
\and Donsub Rim\thanks{Washington University in St. Louis, rim@wustl.edu}
\and Noseong Park\thanks{Yonsei University, noseong@yonsei.ac.kr}
\and Kookjin Lee\thanks{Arizona State University,  kookjin.lee@asu.edu}
}

\date{}

\maketitle







\begin{abstract} \small\baselineskip=9pt
Continuous-time dynamics models, such as neural ordinary differential equations, have enabled the modeling of underlying dynamics in time-series data and accurate forecasting. However, parameterization of dynamics using a neural network makes it difficult for humans to identify causal structures in the data. In consequence, this opaqueness hinders the use of these models in the domains where capturing causal relationships carries the same importance as accurate predictions, \textit{e.g.}, tsunami forecasting. In this paper, we address this challenge by proposing a mechanism for 
mining causal structures from continuous-time models. 
We train models to capture the causal structure by enforcing sparsity in the weights of the input layers of the dynamics models. We first verify the effectiveness of our method in the scenario where the exact causal-structures of time-series are known as \textit{a priori}. We next apply 
our method to a real-world problem, namely tsunami forecasting, where the exact causal-structures are difficult to characterize. Experimental results show that the proposed method is effective in learning physically-consistent causal relationships while achieving high forecasting accuracy.

\vspace{1.0em}
\end{abstract}

\section{Introduction.}
%
%
An emerging paradigm for modeling the underlying dynamics in time-series data 
is to use continuous-time dynamics models, 
such as neural ordinary differential equations (NODEs)~\cite{chen2018neural, rubanova2019latent, dupont2019augmented}.
Widely known as a continuous-depth extension of residual networks~\cite{he2016deep}, 
NODEs have a great fit for data-driven dynamics modeling as they construct models in the form of systems of ODEs.
This new paradigm has enabled breakthroughs in many 
applications, 
such as 
patient status/human activity prediction~\cite{rubanova2019latent}, computational physics problems~\cite{lee2021parameterized,lee2021machine,lee2021structure}, or climate modeling~\cite{hwang2021climate, park2020hurricane}. 

A promising approach to improve the performance of those models 
is to increase the \emph{expressivity} of the neural network
they use to parameterize a system,
\textit{e.g.}, by employing convolutional neural networks 
or by augmenting extra dimension in the state space
~\cite{dupont2019augmented}. 
However, the more complex neural networks are, the more challenging 
humans are to interpret the learned models.
It will be even more problematic for tasks that require interpretable \textit{causality}, 
\textit{e.g.}, a high-consequence event prediction, as in tsunami forecasting.

In this work, we study a mechanism for 
mining the causality in time-series data from continuous-time neural networks trained on it.
Specifically, we ask:
\begin{quote}
\emph{%
    How can we identify causal structures from a continuous-time model?
    How can we promote the model to learn causality in time-series? 
}
\end{quote}
This work focuses on Granger causality~\cite{granger1969investigating},
a common framework that are used to quantify 
the impact of a past event observed on the future evolution of data.
\smallskip

\noindent \textbf{Our contributions.}
\emph{First}, we propose a mechanism for extracting the causal structures 
from the parameters of a continuous-time neural network. 
We adapt the concept of component-wise neural networks, studied in~\cite{tank2021neural}, for continuous-time models. 
We enforce column-wise sparsity to the weights of the input layers so that the impact of the input elements of less contributions (less causal) can be 
small. This is achieved by using a training algorithm, which 
minimizes 
data-matching loss and sparsity-promoting loss and prunes the columns whose weights are smaller than a 
threshold. At the end of the training, the norms of columns corresponding to the important input elements will have high magnitudes, which will make interpretation on causality easier.

\emph{Second}, we test our approach if the learned causal structures match with the known ground-truth causal structures.
To evaluate, 
we train two continuous-time dynamics models, NODEs~\cite{chen2018neural} and neural delay differential equations (NDDEs)~\cite{zhu2020neural}, 
on the data sampled from the Lorenz-96 and Mackey--Grass systems, respectively.
The results demonstrate that 
our mechanism 
precisely extracts the causal structures from the data.

\emph{Third}, we 
further evaluate our approach in tsunami forecasting at the Strait of Juan de Fuca~\cite{melgar2016kinematic},
where we do not know the exact causal relationships in the data. 
We train NDDEs on the tsunami 
dataset~\cite{melgar_2016_59943} by using the proposed training algorithm.
We 
achieve comparable accuracy with the prior work~\cite{liu2021comparison} and effective in predicting the highest peaks of sea-surface elevations at locations near highly-populated areas.
Furthermore, we show that one can approximately capture the physically-consistent causal relationships between a tsunami event and tide observed before, which agrees with the domain expert's interpretation.

This result suggests that 
we can accurately model the underlying dynamics in time-series data
using continuous-time dynamics models while capturing the causality.
Future work will explore adopting diverse causality definitions
and apply to various 
applications.

\section{Preliminaries}
\label{sec:prelim}


\subsection{Neural Ordinary Differential Equations}\hspace*{\fill}
\label{subsec:nodes}

\noindent
Neural ordinary differential equations (NODEs)~\cite{chen2018neural}
are a family of deep 
neural networks 
that parameterize the time-continuous dynamics of hidden states in the data 
as a system of ODEs using a neural network:
\begin{equation}\label{eq:node}
    \dv{\LatentState}{t} = \Velocity_{\NNParams}(\LatentState, t),
\end{equation}
where $\LatentState(t) \in \mathbb{R}^{\NumLatentState}$ denotes a time-continuous hidden state, 
$\Velocity_{\NNParams}$ denotes a velocity function, parameterized by a 
neural network whose 
parameters are denoted as $\NNParams$. 
A typical parameterization of $\Velocity_{\NNParams}$ is 
a multi-layer perceptron (MLP) 
$\NNParams = \{ (\NNWeight^{\ell}, \NNBias^{\ell}) \}_{\ell=1}^{}$. 
$\NNWeight^{\ell}$ and $\NNBias^{\ell}$ are weights and biases of the $\ell$-th layer, respectively.

The forward pass of 
NODEs is equivalent to solving an initial value problem 
via a black-box ODE solver. 
Given an initial condition $\LatentState_0$ and $\Velocity_{\NNParams}$, it computes: 
\begin{equation}\label{eq:ivp}
    \LatentState({t_0}), \LatentState(t_1) = \text{ODESolve}(\LatentState_0, \Velocity_{\NNParams}, \{ t_0,t_1\}). 
\end{equation}

\subsection{Neural Delay Differential Equations}\hspace*{\fill}
\label{subsec:nddes}

\noindent 
Although NODEs serve as a great means for data-driven dynamics modeling, they exhibit several limitations. 
One obvious limitation in the context of causality modeling is that NODEs take only the current state of the input variables $\LatentState(t)$ as shown in Eq.~\eqref{eq:node} and, thus, are not suitable for capturing delayed causal effects.  

To resolve this issue, we employ the computational formalism 
provided by neural delay differential equations (NDDEs)~\cite{zhu2020neural}, an extension of NODEs, which takes extra input variables $\LatentState_{\leq \tau}(t)$ as follows:
\begin{equation}
    \dv{\LatentState}{t} = \Velocity_{\NNParams}(\LatentState(t), \LatentState_{\leq \tau}(t),t),
\end{equation}
where $\LatentState_{\leq \tau}(t) = \{\LatentState (t - \gamma): \gamma \in [0, \tau]\}$ denotes the trajectory of the solution in the past up to time $t - \tau$.
To avoid numerical challenges in handling a continuous form of delay, 
we choose to consider 
a discrete form of delay:
\begin{equation} \label{eq:discrete_time_delay}
    \dv{\LatentState}{t} = \Velocity_{\NNParams}(\LatentState(t),\LatentState(t-\tau_1),\ldots, \LatentState(t-\tau_m),t).
\end{equation}

\subsection{Neural Granger Causality}\hspace*{\fill}
\label{subsec:neural-granger-causality}

\noindent
Granger causality~\cite{granger1969investigating} has been a common choice for discovering a structure that quantifies the extent to which one time series affects 
predicting the future evolution of another time series. When the linear dynamics model is considered, such as a vector autoregressive model, where the dynamics 
are simply represented as a linear transformation, \textit{e.g.}, $\LatentState_t = A \LatentState_{t-1}$. 
Granger causality can be identified by the structure of the matrix $A$. 
If the $i$-th column of $j$-th row is zero, it can be interpreted that the $i$-th series does not Granger cause the $j$-th series. For time series that evolves according to general nonlinear dynamics, however, it is often challenging to embed such structure in a nonlinear dynamics model. 

Recent work~\cite{tank2021neural} 
proposed to adapt neural network architectures in a way that the Granger causal interactions (or neural Granger causality) are estimated while performing data-driven dynamics modeling of nonlinear systems.
They presented component-wise MLPs or component-wise recurrent neural networks, where the individual components of the output are constructed as separate neural networks and, thus, the effects of input on individual output series can be captured. To promote the sparse causality structure for better interpretability, 
their method imposes a penalty on the input layers of each component neural networks. 
In this work, we translate their causality learning frameworks into NODEs and NDDEs.

\section{Mining Causality in Time-series Data}
\label{sec:method}

We now define neural Granger causality in the context of continuous-time dynamics models.
We then propose a mechanism for mining the causality in time-series via training continuous-time dynamics models on it. 

\subsection{Neural Granger Causality in\\ \protect\hspace*{2.0em} Continuous-time Dynamics Models}\hspace*{\fill}
\label{subsec:ngc-in-cont-models}

\noindent 
As a multi-variate time-series, in continuous-time models, 
represented as an implicit form, the rate of changes of the variables with respect to time $t$,
we define neural Granger causality in continuous-time models as follows:

\begin{Definition}\label{def:ngc}
Let us call the time-series $z_j$ with delay $\tau_k$ as the $(j,k)$-th delayed time-series $z_{j, k} (t) = z_j (t - \tau_k)$. We say $(j, k)$-th delayed time series $z_{j,k}$ is \emph{Granger non-causal in the system \eqref{eq:discrete_time_delay}} with respect to the $i$-th time-series $z_i$, if 
$\forall (\LatentStateSymb_{1, 1} , ... , \LatentStateSymb_{j,k}, ... , z_{n_z, m})$ and 
$\forall {\LatentStateSymb'}_{j, k} \neq \LatentStateSymb_{j, k}$,
\begin{equation}
\resizebox{.9\hsize}{!}{
    $\VelocitySymb_i(\LatentStateSymb_{1, 1} , ... , \LatentStateSymb_{j,k}, ... , z_{n_z, m}) = \VelocitySymb_i(\LatentStateSymb_{1, 1} , ... , \LatentStateSymb'_{j,k}, ... , \LatentStateSymb_{n_z, m}),$
    }
\end{equation}
where $\VelocitySymb_i$ is the $i$th element of the velocity function $\Velocity$.
\end{Definition}
Note that, for NODEs, Def.~\ref{def:ngc} can be reduced to the case with no delayed variables.

\subsection{Mining Causal Structures in NODEs}\hspace*{\fill}

\begin{figure}[!ht]
    \centering
    \subfloat[][]{
     \includegraphics[scale=1]{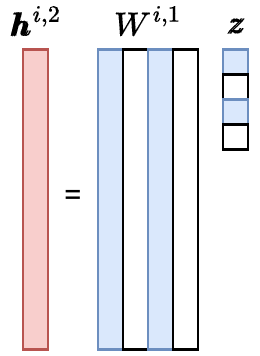} \label{fig:input_layer_nondelay}}
     \hspace{5mm}
    \subfloat[][]{ 
    \includegraphics[scale=1]{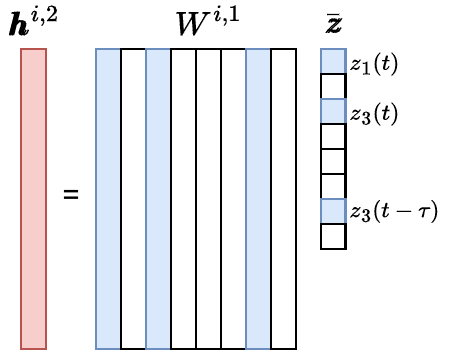} \label{fig:input_layer_delay}}
    \caption{An illustrative operation of the input layer.}
    \label{fig:input_layer}
\end{figure}

\paragraph{Identifying causal structures.}
Inspired by the adaptation proposed by Tank~\textit{et al.}~\cite{tank2021neural}, 
we first construct a component-wise neural network architecture: 
\begin{equation}
    \Velocity_{\NNParams}(\LatentState) = \begin{bmatrix} \VelocitySymb_{1}(\LatentState;\NNParams_1)\\
    \vdots\\
    \VelocitySymb_{\NumLatentState}(\LatentState;\NNParams_{\NumLatentState})\\
    \end{bmatrix} \in \mathbb{R}^{\NumLatentState},
\end{equation}
where $\VelocitySymb_{i}(\LatentState;\NNParams_i) \in \mathbb{R}$ denotes the $i$th element of the function, i.e., $\dv{\LatentState_i}{t} = \VelocitySymb_{i}$. Each element of the velocity function is parameterized as an MLP; 
the $l$th layer of $\VelocitySymb_i$ is represented as $\HiddenState^{i,l+1} = \NonlinearActivation (\NNWeight^{i,l} \HiddenState^{i,l} + \NNBias^{i,l})$. As the highly entangled representation of the internal layers does not allow enforcing Granger causality, the disentanglement has to be captured in the input layer. For example, if the $j$th column of the first layer weight, $\NNWeight^{i,1}$, of $\VelocitySymb_i$ consists of zeros, time series $j$ is Granger non-causal for time series $i$. Figure~\ref{fig:input_layer} depicts an illustrative example, where the $i$th time series has Granger causality from time series 1 and 3 (blue). The Granger non-causal elements, i.e., series 2 and 4 (white), can be filtered out by zeroing the second and the fourth columns of $\NNWeight^{i,l}$. 

\paragraph{Learning causality via sparsity-promoting loss and pruning.}
Training of NODEs are performed by minimizing the discrepancy between their predictions 
and the ground-truth data, $\sum_{i=1}^{\nTrain} L(z(t_i), \tilde z(t_i))$, where $L$ is a certain measure of discrepancy, \textit{e.g.}, $\ell^2$-norm. To promote the sparsity in the weight matrices $\{\NNWeight^{i,1}\}_{i=1}^{\NumLatentState}$, 
we add a column-wise penalty to the training loss: 
\begin{equation}\label{eq:loss}
    \sum_{i=1}^{\nTrain} L(z(t_i), \tilde z(t_i)) + \alpha \sum_{k=1}^{\NumLatentState} \sum_{l=1}^{\NumLatentState} \| [\NNWeight]_l^{k,1} \|_2,
\end{equation}
where $[\NNWeight]_l$ denotes the $l$-th column of $\NNWeight$, and $\|\cdot\|_2$ is the $\ell_2$-norm of a vector. 
$\alpha$ is the hyper-parameter defining the importance between the two terms.
By enforcing the column-wise norms to be minimized, columns associated with input elements with non-significant contributions are forced to have small numerical values.

In addition to the sparsity-promoting loss, we introduce a magnitude-based pruning method to capture the causality structure more explicitly. As the training proceeds, we prune a column whose $\ell_2$-norm becomes smaller than a certain threshold, $\prunethreshold$: 
\begin{equation}\label{eq:prune}
    [W]_l^{k,1} = \bm{0} \quad \text{if} \quad  \| [W]_l^{k,1} \|_2 \leq \prunethreshold,
\end{equation}
where $\bm{0}$ is a vector consisting of zeros.

\subsection{Mining Causal Structures in NDDEs}\hspace*{\fill}
\label{subsec:mine-causality-in-nddes}

\noindent
The velocity function of NDDEs 
has the same structure with NODEs:
\begin{equation}
    \Velocity_{\NNParams}(\bar{\LatentState}(t)) = \begin{bmatrix} \VelocitySymb_{1}(\LatentState(t),\LatentState(t-\tau_1),\ldots,\LatentState(t-\tau_m) ;\NNParams_1)\\
    \vdots\\
    \VelocitySymb_{\NumLatentState}(\LatentState(t),\LatentState(t-\tau_1),\ldots,\LatentState(t-\tau_m);\NNParams_{\NumLatentState})\\
    \end{bmatrix},
\end{equation}
where 
$\VelocitySymb_{i}\in \mathbb{R}$ denotes the $i$-th element of the velocity function, \textit{i.e.}, $\dv{\LatentState_i}{t} = \VelocitySymb_{i}$. 

The input layer of the $i$-th velocity element, $\VelocitySymb_i$, is represented as 
$\NNWeight^{i,1} \bar{\LatentState} + \NNBias^{i,1}$, where $\bar{\LatentState}(t)$ is a vertical concatenation of all delayed variables:
\begin{equation}
    \begin{split}
    \bar{\LatentState}(t) &=[\LatentState(t)\Transpose, \LatentState(t-\tau_1)\Transpose,\ldots,\LatentState(t-\tau_m)\Transpose]     \\
    &= [\LatentStateSymb_1(t),\ldots,\LatentStateSymb_{\NumLatentState}(t), \LatentStateSymb_1(t-\tau_1),\ldots,\LatentStateSymb_{1}(t-\tau_m),\ldots]. 
    \end{split}
\end{equation}
With the same reason described above in identifying causal structure in NODEs, if the $j$th column of the first layer weight, $\NNWeight^{i,1}$, the $j$th element in $\bar{\LatentState}(t)$ is Granger non-causal for time series $i$.

\paragraph{Learning causality and automatic lag detection.}
To identify causal structures of the multivariate time-series over the course of the training, the sparsity promoting loss is minimized (as in Eq.~\eqref{eq:loss}) along with the pruning algorithm (as in Eq.~\eqref{eq:prune}).
Via the sparsity-promoting loss and the pruning method, it is expected that the entries of $\bar{\bm{\LatentState}}$, that are Granger non-causal to specific time-series, are expected to pruned and, thus, the lags where Granger causal effects exist can be detected automatically. 

\subsection{Putting All Together}\hspace*{\fill}
\label{subsec:all-in-one}

\noindent
In the actual implementation, we employ the standard mini-batching and a variant of stochastic gradient descent (SGD) to train the neural network architectures described in the previous subsections. With the sparsity promoting penalty and the pruning, the entire training process shares the commonality with the training algorithm proposed in~\cite{lee2021structure}, 
shown in Algorithm~\ref{alg:train}.
\begin{algorithm} 
\caption{NODEs/NDDEs Training}\label{alg:train}
\begin{algorithmic}
\State Initialize $\NNParams$
\For{$(i = 0;\ i <n_{\max} ;\ i = i + 1)$}
    \State Sample $n_{\text{batch}}$ trajectories randomly from $\mathcal D_{\text{train}}$ 
    \State Sample initial points randomly from the sampled trajectories: 
     $\bar{\LatentState}^{r}({s(r)})$, $s(r)\in[0,\ldots,m-\ell_{\text{batch}}-1]$
    for $r=1,\ldots,n_{\text{batch}}$
    \State 
    $\tilde{\LatentState}(t_1),\ldots,\tilde{\LatentState}(t_m)\!\! =\!\! \mathrm{ODESolve}(\bar{\LatentState}^{r}_{s(r)}, \Velocity_{\NNParams},t_1, \ldots, t_{m})$,
    for $r=1,\ldots,n_{\text{batch}}$
    \State 
    Compute the loss (Eq. \eqref{eq:loss}) 
    \State Update $\NNParams$ via SGD
    \State Prune $\NNParams$ based on the magnitude (Eq. \eqref{eq:prune})
\EndFor
\end{algorithmic}
\end{algorithm}
We implement our method in \textsc{Python} using a deep learning framework, \textsc{PyTorch}~\cite{paszke2019pytorch}. For the NODEs/NDDEs capability,
we use the \textsc{TorchDiffeq} library~\cite{chen2018neural}. 
We will use the notation $\TimeSeries$ to distinguish the time-stepped-series as opposed to the continuous series $\LatentState$.

\section{Mining Known Causal Structures}

In this section, we showcase the effectiveness of the proposed method with two canonical benchmark problems: the Lorenz-96 system and the Mackey--Glass equation. The Lorenz-96 system has been an important testbed for climate modeling. We choose this system to 
show the causality learning in the context of NODEs (with the assumption that there is no delayed effect). The Mackey--Glass (MG) equation is another chaotic system, describing the healthy and pathological behaviour in certain biological contexts (\textit{e.g.}, blood cells). We use MG to 
demonstrate the causality learning with delayed variables in the context of NDDEs.


\subsection{Benchmark 1: The Lorenz-96 System}\hspace*{\fill}
\label{subsec:lorenz96}

\noindent
The 
Lorenz-96 system \cite{lorenz1996predictability} is given by the equations: 
\begin{equation}
    \dv{x_i}{t} =  (x_{i+1}-x_{i-2})x_{i-1} - x_i + F,
\end{equation}
where $i=1,\ldots,N$ with $x_{-1} = x_{N-1}$, $x_{0} = x_{N}$, and $x_{1} = x_{N+1}$, and $F$ denotes the forcing term. 

\noindent \textbf{Setup.}
We consider the system with $N=6$ variables. 
For modeling, 
we use the component-wise NODE where each MLP, $f_i$, has 4 layers with 100 neurons. 
For the nonlinearity, we use the hyperbolic tangent (Tanh).
The detailed experimental setup is in Appendix.

\begin{figure}[!ht]
    \centering
     \includegraphics[width=.9\linewidth]{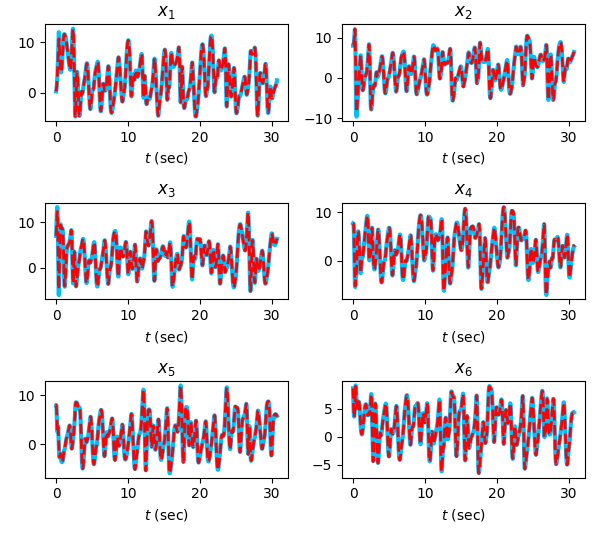} 
    \caption{[Lorenz-96] The ground-truth trajectories (solid blue) and the predicted trajectories computed from the learned model (dashed red).}
    \label{fig:lorenz_pred}
    \vspace{-1.0em}
\end{figure}

\begin{figure}[!t]
    \centering
    \subfloat[][$N=6$]{
    \hspace{-5.mm}
     \includegraphics[scale=.55]{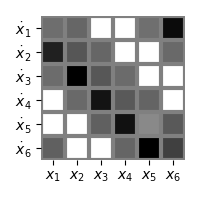} }
     \subfloat[][$N=8$]{
     \hspace{-2.5mm}
     \includegraphics[scale=.55]{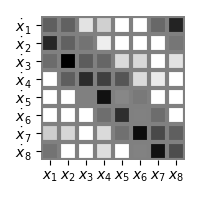} }
     \subfloat[][$N=10$]{
    \hspace{-2.5mm}
     \includegraphics[scale=.58]{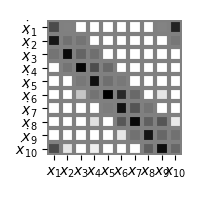} }
    \caption{[Lorenz-96] Identified causality structure for the systems with $N=6, 8, 10$.}
    \label{fig:lorenz_classif}
\end{figure}

\noindent \textbf{Results.} 
Figure~\ref{fig:lorenz_pred} depicts the ground-truth trajectories and the predicted trajectories and we observe that 
the predictions match well with the ground-truth trajectories. Figure~\ref{fig:lorenz_classif} reveals the learned causality structure by showing the the  magnitude of the column norms of the weight matrices, $\{\NNWeight^{i,1}\}$, in the input layer;  the magnitude is normalized to have 1 as the maximum value (black) and 0 as the minimum value (white). The  horizontal entries (\textit{i.e.}, $\{x_i\}$) with darker colors (close to the black color) can be considered as the ones with the significant contributions to the vertical entries (\textit{i.e.}, $\{\dot{x}_i\}$). Figure shows that the models have learned that $\dot{x}_i$ is Granger causal with $x_{i-2}, x_{i-1}, x_i, x_{i+1}$ and non-Granger causal with other entries. 

\subsection{Benchmark 2: The Mackey--Glass System}\hspace*{\fill}
\label{subsec:mg}

\noindent
The next benchmark problem, the Mackey--Glass (MG) system \cite{mackey1977oscillation} is given by the equation: 
\begin{equation}
    \dv{x}{t} = -b x(t) + a \frac{x(t-\tau)}{1 + x(t-\tau)^c},
\end{equation}
where $a,b$ and $c$ denote the ODE parameters and $\tau$ denotes the lag. The values of $x(t)$ for $t\leq 0$ is defined by the initial function $\phi(t) = .5$. 

\noindent \textbf{Setup.}
For parameterizing NDDEs, we model the right-hand side of the NDDE as an MLP that takes $m=10$ candidate delayed variables along with the current variable as an input, such that 
\begin{equation}
    \bar{\bm{x}}(t) = [x(t), x(t-\tau_1),\ldots,x(t-\tau_{10})] \in \mathbb{R}^{11}, 
\end{equation}
where $\tau_i = i$ (seconds), for $i=1,\ldots,10$. 
We consider an MLP consisting of 4 layers with 25 neurons and use the Tanh fuction for the nonlinearity.
We refer readers to Appendix for the detailed experimental setup.

\begin{figure}[!ht]
    \centering
     \includegraphics[scale=.5]{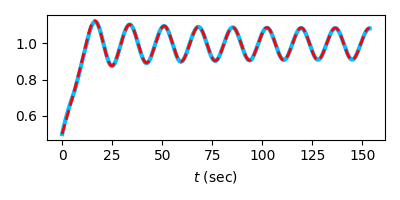} 
    \caption{[Mackey--Glass] The ground-truth trajectory (solid blue) and the predicted trajectory computed from the learned model (dashed red).}
    \label{fig:mg_pred}
\end{figure}

\begin{figure}[!ht]
    \centering
     \includegraphics[scale=.85]{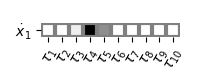} 
    \caption{[Mackey--Glass] Identified causality structure for the systems with 10 delayed variables.}
    \label{fig:mg_classif}
\end{figure}

\noindent \textbf{Results.}
Figure~\ref{fig:mg_pred} depicts the ground-truth trajectory and the predicted trajectory computed from the learned NDDE model. We repeated the same training for five times with different initializations. Although the mean and the two standard deviation are plotted in Figure~\ref{fig:mg_pred}, each prediction appear almost identical. Figure~\ref{fig:mg_classif} depicts the learned causality structure. 
The values are normalized to lie between [0, 1]. We can observe that the most significant contributes are from $x(t-{\tau_4})$ and $x(t-{\tau_5})$, meaning that the delayed effect appears within a 4$\sim$5-seconds window, which is slightly off to the ground truth delay (i.e., $\tau=5$). We believe this is due to the difference between two numerical solvers that we use for generating ground-truth trajectory and simulating data-driven models. 

\section{An Application to Tsunami Forecasting}

Here, we further investigate the forecasting and causality mining capabilities of the NDDEs equipped with the learning algorithm proposed in Sec.~\ref{sec:method}. 
We examine them with 
a dataset of tsunami realizations used for developing tsunami forecasting models~\cite{liu2021comparison}. There exist only a handful of real earthquake events with large magnitudes, so synthetic tsunami data has been generated using numerical tsunami simulations that model the physics. In particular, the tsunami wave propagation modeled by nonlinear partial differential equations~\cite{leveque_george_berger_2011}. The simulation is initiated by an incoming tsunami wave, which interacts nonlinear with the topography and reflecting waves. This particular setup makes the dataset ideal for interpreting the causality results.

\subsection{Tsunami Dataset}\hspace*{\fill}
\label{subsec:tsunami-dataset}

\noindent
The dataset was created from a set of 1300 synthetic Cascadia Subduction Zone (CSZ) earthquake events ranging in magnitude from Mw 7.8 to 9.3, as described \cite{melgar2016kinematic} and made available in~\cite{melgar_2016_59943}.  These were generated from methods proposed in \cite{leveque2016generating} using the MudPy software \cite{Melgar2020}. The resulting seafloor deformation was then used as initial conditions for tsunami wave propagation implemented in \cite{clawpack}. 

\begin{figure}[!h]
    \centering
     \includegraphics[width=\columnwidth]{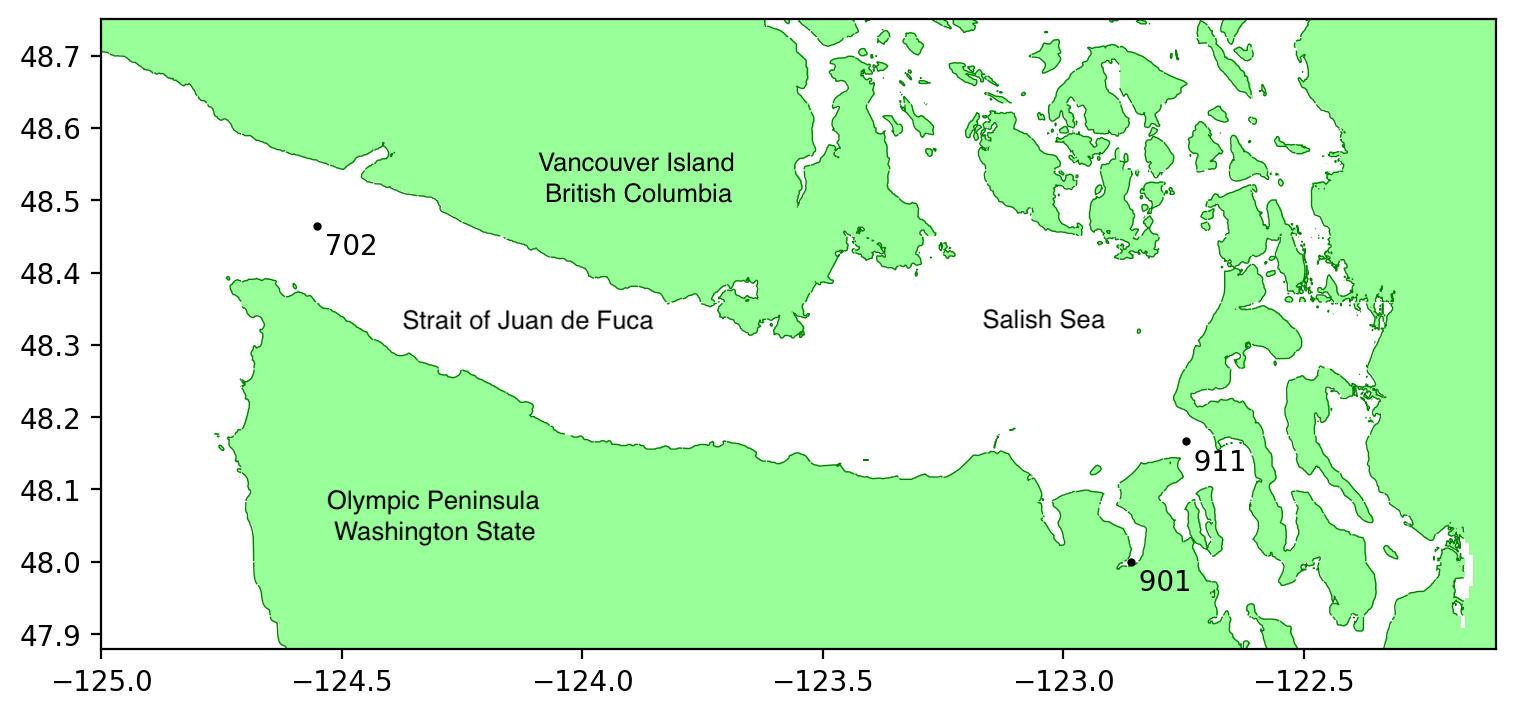} 
    \caption{[Tsunami] The geography of the tsunami simulation depicting the Strait of Juan de Fuca, and the location of the Gauges 702, 901, and 911.}
    \label{fig:tsunami_gauge_loc}
\end{figure}

\begin{figure}[t]
    \centering
     \includegraphics[scale=.5]{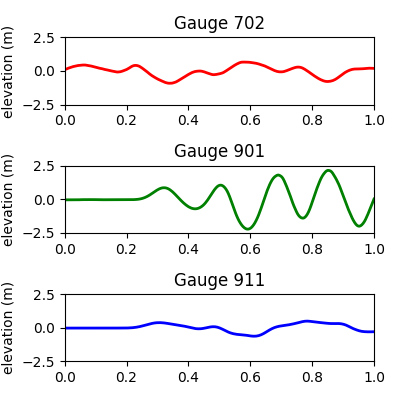} 
    \caption{[Tsunami] An example plots of Surface elevation measured at Gauge 702, 901, and 911.}
    \label{fig:tsunami_ts}
\end{figure}

The tsunami data for one earthquake event contains tri-variate time-series data 
with the variables $(\TimeSeriesSevenOTwo, \TimeSeriesNineOOne,\TimeSeriesNineOneOne)$. The time series has duration of 5-hours and is interpolated at 256 uniformly spaced points on the time-grid. This time-series data corresponds to gauge readings for the synthetic tsunami entering the Strait of Juan de Fuca (SJdF). Each variable corresponds to different geological locations denoted by the number designations 702, 901, and 911, as shown in Figure~\ref{fig:tsunami_gauge_loc}. Gauge 702 is located at the entrance of SJDF, and the other two gauges are located further inside from the entrance: Gauge 901 is located in Discovery Bay, and Gauge 911 is located in the middle of Admiralty Inlet. Figure~\ref{fig:tsunami_ts} shows an example of the surface elevation time-series measured at the three gauges.

Liu \textit{et al.}~\cite{liu2021comparison} considered tsunamis originating from hypothetical megathurst earthquakes in the CSZ that reach the Puget Sound. Since 
SJdF 
is the only path for the tsunamis to reach  high-population areas in the Sound (see Figure~\ref{fig:tsunami_gauge_loc}), the authors hypothesized whether observing the tsunami near the entrance of the strait at Gauge 702 could be used to 
forecast 
its amplitude at Gauge 901 and 
911.




Among the 1300 time-series instances of tsunami, the events that with negligible tsunami 
in the region of our interest were discarded; event with tsunami amplitudes less than 0.1m at Gauge 702 or 0.5m at Gauge 901 were removed from the dataset. This leads to the decrease in the number of time-series instances in the dataset from 1300 to 959.
%
We then split the remaining data into 80/5/15 for the train/validation/test set.

%
%

\subsection{Model Architecture}\hspace*{\fill}
\label{subsec:architecture}

\noindent
We employ an NDDE, parameterized 
by using an MLP, 
that takes 
the three variables as an input and the their delayed versions as follows:
\begin{equation}
    \begin{split}
        \bar{\TimeSeries} (t) = [&\TimeSeriesSevenOTwo(t), \TimeSeriesSevenOTwo(t-\tau_1),\ldots, \TimeSeriesSevenOTwo(t-\tau_m),\\
        &\TimeSeriesNineOOne(t),\TimeSeriesNineOOne(t-\tau_1),\ldots,\TimeSeriesNineOOne(t-\tau_m),\\
        &\TimeSeriesNineOneOne(t), \TimeSeriesNineOneOne(t-\tau_1), \ldots, \TimeSeriesNineOneOne(t-\tau_m)],
    \end{split}
\end{equation}
and outputs the time-derivatives of the three variables 
\begin{equation}
\dot{{\TimeSeries}} = \left[\frac{\mathrm d \TimeSeriesSevenOTwo}{\mathrm d t}, \frac{\mathrm d \TimeSeriesNineOOne}{\mathrm d t}, \frac{\mathrm d\TimeSeriesNineOneOne}{\mathrm d t}\right].   
\end{equation}

We 
consider an MLP with 4 layers with 100 neurons and Tanh nonlinearity for each output element. As in the empirical study \cite{gusak2020towards} of applying different normalization techniques to NODEs, we 
examined several combinations of normalization techniques, including layer normalization (LN) \cite{ba2016layer}, weight normalization (WN) \cite{salimans2016weight}, and spectral normalization \cite{miyato2018spectral}. 
We found the best working 
configuration for this particular case study is
\begin{equation}
    [\text{Linear} \rightarrow \text{WN} \rightarrow \text{LN} \rightarrow \text{Tanh}] \times 4.
\end{equation}

We 
ran a hyperparameter 
search with 9 combinations out of the weight penalty $\alpha$ in $\{0.1,0.01,0.001\}$ and the pruning threshold $\prunethreshold$ in $\{0.1,0.01,0.001\}$. 
We found that $\alpha=\prunethreshold=0.01$ yields the best result. 
Promoting 
strong sparsity (larger $\alpha$ and $\prunethreshold$) or 
weak sparsity (smaller $\alpha$ and $\prunethreshold$) 
resulted in degradation in forecasting accuracy. 
We use this setting for our experiments.

\subsection{Tsunami Forecasting Accuracy}\hspace*{\fill}
\label{subsec:tsunami-results}

\begin{figure}[!h]
    \centering
     \includegraphics[scale=.6]{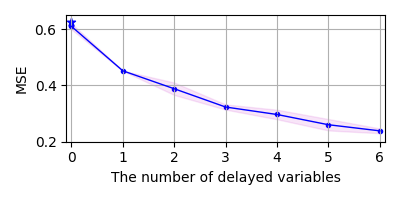} 
    \caption{[Tsunami] Averaged mean-squared errors (MSE) for varying $m$ (the number of delayed variables). Star marker indicates the MSE of ANODE results.}
    \label{fig:tsunami_mse}
\end{figure}

\begin{figure*}[!h]
    \centering
    \subfloat[][NODEs]{
     \includegraphics[width=0.6\columnwidth]{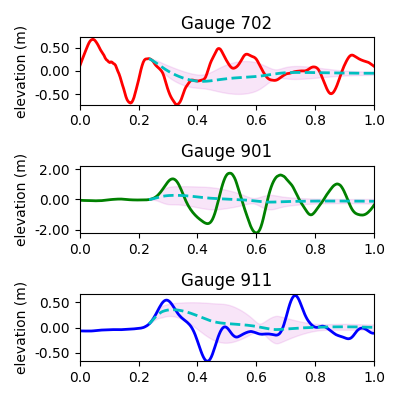} \label{fig:tsunami_node}}\hspace{2.5mm}
     \subfloat[][NDDEs (with $m=3$)]{
     \includegraphics[width=0.6\columnwidth]{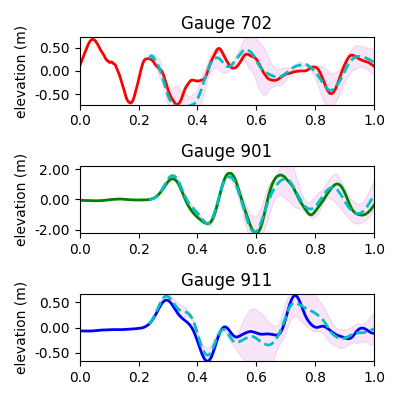} \label{fig:tsunami_ndde3}}\hspace{2.5mm}
     \subfloat[][NDDEs (with $m=6$)]{
     \includegraphics[width=0.6\columnwidth]{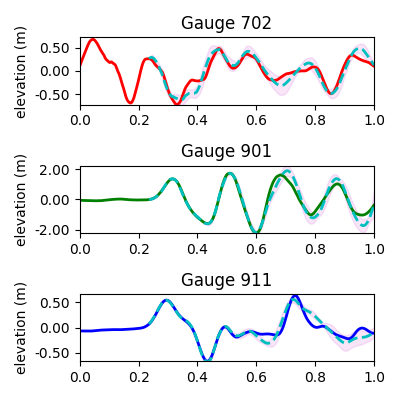} \label{fig:tsunami_ndde}}
    \caption{[Tsunami] An example of prediction results of using the trained NODEs and NDDEs.}
    \label{fig:tsunami_node_ndde}
    \vspace{-1.0em}
\end{figure*}

\noindent
To evaluate,
we train three continuous-time dynamics models, \textit{i.e.}, NODEs, ANODEs, and NDDEs with varying number of delayed variables, on the tsunami dataset and measure those models' forecasting accuracy as the mean-squared errors (MSE).
For each model, 
we repeat the experiments 5 times with different initialization of model parameters. In Figure~\ref{fig:tsunami_mse}, 
we report the average 
mean-squared errors (blue line) and the area covered by the two standard deviation (magenta area).
\paragraph{Results with NODEs.} Figure~\ref{fig:tsunami_node} depicts an example instance of the elevation time series in the test set and the prediction made by solving an IVP given the trained NODE model and the initial condition obtained from the test set. 
We report the mean of the predictions (cyan dashed line) and the two standard deviations (magenta area). 
The trained NODEs 
can only produce simple trajectories, which do not forecast the oscillatory behavior 
in the data and fail to match the peak values and the locations 
in the time series. We also test ANODEs, but ANODEs do not 
provide 
significantly different results than NODEs 
(see 
Appendix).
 
\paragraph{Results with NDDEs.} Next, we 
test the NDDEs with varying number of delayed variables (\textit{i.e.}, $m$ in $\{1,2,3,4,5,6\}$) where $\tau_i = 12i $ (min). As depicted in Figure~\ref{fig:tsunami_mse}, the increase in the number of delayed variables results in very models in terms of prediction accuracy measured in mean-squared errors. 

\begin{figure}[!h]
    \subfloat[][Gauge 901 ($m=3$)]{
     \includegraphics[width=0.48\columnwidth]{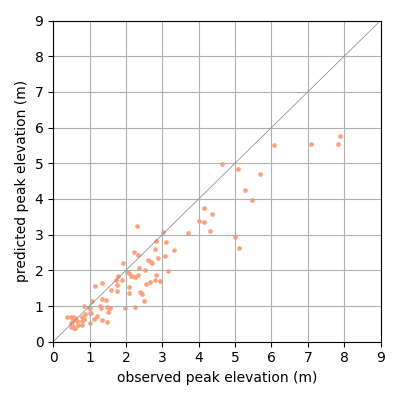}}
     \subfloat[][Gauge 901 ($m=6$)]{
     \includegraphics[width=0.48\columnwidth]{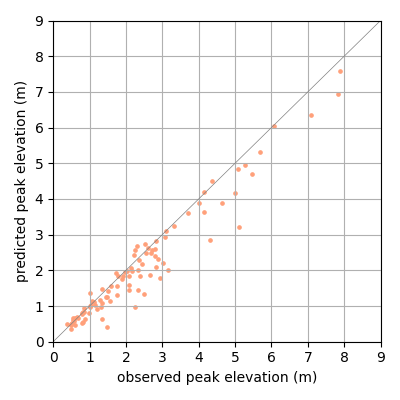}}\\
     \subfloat[][Gauge 911 ($m=3$)]{
     \includegraphics[width=0.48\columnwidth]{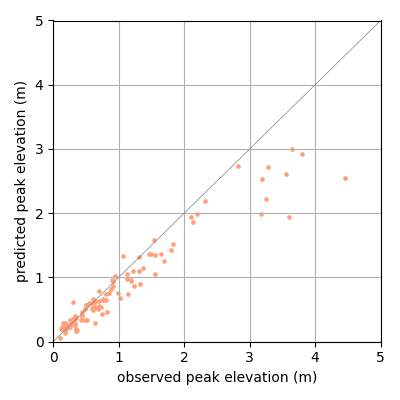}}
     \subfloat[][Gauge 911 ($m=6$)]{
     \includegraphics[width=0.48\columnwidth]{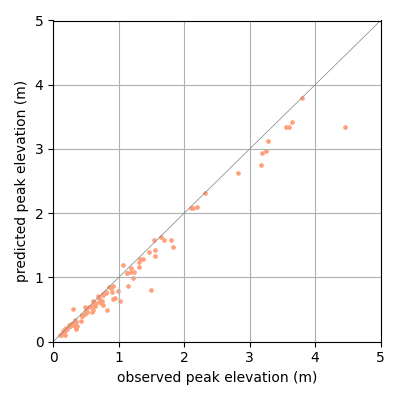}}
    \caption{[Tsunami] Observed peak locations (horizontal axis) and predicted peak locations (vertical axis) for elevation time-series measured at Gauge 901 and 911. 
    We use the data from the test-set.}
    \label{fig:tsunami_peaks}
\end{figure}

The figure also shows the elevation time series for the same test data 
considered in the 
experiments with NODEs.
We plot the mean of the predictions as cyan dashed lines and the two standard deviations (as magenta areas). 
We use the same experimental procedure for forecasting (\textit{i.e.}, solving IVP with the learned NDDEs). 
As opposed to the predictions made by NODEs, 
the predictions made by NDDEs match well with the ground-truth trajectories 
and 
capture the peak values and locations more accurately.
Figure~\ref{fig:tsunami_peaks} 
show the predicted peak values more aligned with the ground-truth trajectories
at Gauge 901 and 
911. Two sets of the predictions made separately by NDDEs with $m=3$ and NDDEs with $m=6$ are depicted. 
The NDDEs with larger $m$ provide better predictions for peak values.

\begin{figure*}[!ht]
    \centering
    \includegraphics[scale=.65]{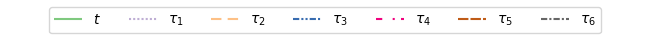} \\
    \vspace{-1.0em}
    \subfloat[][ $\dv{\TimeSeriesSevenOTwo}{t}$]{
     \includegraphics[width=.65\columnwidth]{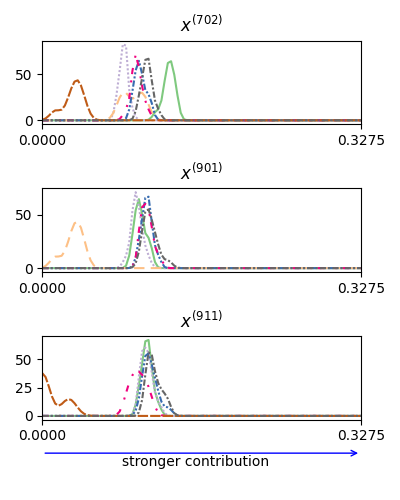} \label{fig:causality702}}
    \subfloat[][$\dv{\TimeSeriesNineOOne}{t}$]{
     \includegraphics[width=.65\columnwidth]{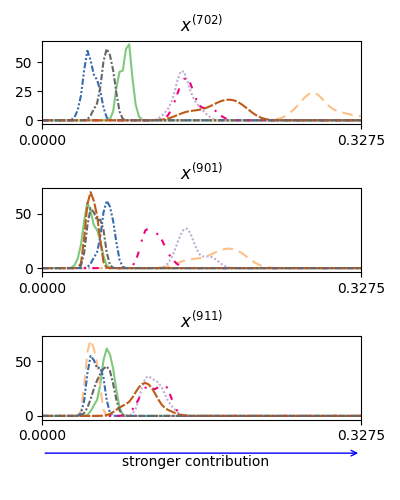} \label{fig:causality901}}
     \subfloat[][$\dv{\TimeSeriesNineOneOne}{t}$]{
     \includegraphics[width=.65\columnwidth]{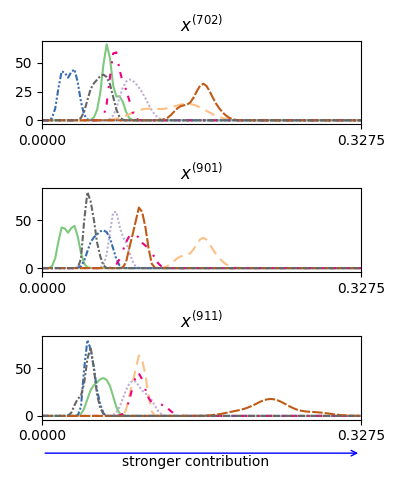} \label{fig:causality903}}
    \caption{[Tsunami] 
    Causal structures of NDDE models with $m=6$: the contributions from  $\TimeSeriesSevenOTwo$ (top), $\TimeSeriesNineOOne$ (middle), and $\TimeSeriesNineOneOne$ (bottom) to $\DotTimeSeriesSevenOTwo$ (a, left), $\DotTimeSeriesNineOOne$ (b, middle), and $\DotTimeSeriesNineOneOne$ (c, right).}
    \label{fig:tsunami_causality}
    \vspace{-1.0em}
\end{figure*}

\subsection{Mined Causal Structures}\hspace*{\fill}
\label{subsec:tsunami-causal-structure}

\noindent
In Figure~\ref{fig:tsunami_causality}, we report the learned causal structure. The magnitude of the column norms of the weight matrices $\{\NNWeight^{i,1} \}$ in the input layer are again considered as the amount of contributions made by the corresponding input variables. We repeat the same experiments 15 times with 
different initialization, obtain the column norms from each run, and then apply the kernel density estimation based on Scott's factor \cite{scott2015multivariate} to the collected column norms.

We first focus on the plots on the diagonal of the 3$\times$3 plot matrix. Gauge 702 has only small contributions from the delayed variables. The gauge observes the incoming wave and there is little reflections coming from the interior of the Sound, so it is expected that there are only small contributions the causality result. Gauges 901 and 911 have higher contributions since they do possess causal behavior coming from the reflections caused by the narrowing strait. This is especially pronounced for 901 which sits in Discovery bay and experiences significant sloshing due to the local topography (see Figures~\ref{fig:tsunami_ts} and \ref{fig:tsunami_node_ndde}). Next, we observe that the upper diagonal plots reveal larger contributions where as the lower-diagonal plots do not. This indicates that the delayed time-series of Gauge 702 causes those of Gauge 901 and 911 in a significant way, and that delayed time-series at Gauge 901 causes Gauge 911. The significant contributions agree well with the spatio-temporal progression of the physical wave itself. 

The most significant delayed variables from Gauge 702 time-series in predicting Gauge 901 are the delays $\tau_2$ and $\tau_5$, which correspond roughly to 24 minutes and 60 minutes of delay. Thus 30-60 minutes of the time-series from 702 is required to predict the time-series at Gauge 911, which agrees with the prediction results from \cite{liu2021comparison}: while 30 minutes of time-series from Gauge 702 was sufficient to predict the wave height at Gauge 901, using 60 minutes improved the result significantly.

\section{Related Work}
\label{sec:related-work}

NDDEs have been studied mostly for general machine learning 
benchmarks, \textit{e.g.}, image classification~\cite{zhu2020neural} or canonical ODE example problems~\cite{holt2022neural, widmann2022delaydiffeq}. However, to our knowledge, this is the first work investigate the application of NDDEs in real-world time-series data.

In the field of data mining, there are several tsunami or earthquake-related studies, predicting human mobility behaviour~\cite{song2014prediction} and infrastructure damage~\cite{priya2020endea} after natural disaster, evacuation planning with pre-disaster web behavior~\cite{yabe2019predicting}, forecasting earthquake signals~\cite{huang2020crowdquake, zhang2019aftershock}, to name a few. A study closest to ours is in~\cite{piatkowski2012spatio}, where a spatio-temporal model for Tsunami forecasting, based on buoys in the Pacific Ocean, has developed. However, their prediction scale is in much longer time scale (3-12 hours) than ours. More importantly, the study only focuses on model performance on forecasting without considering underlying causal structures of the time-series.

\section{Conclusion}
\label{sec:conclusion}

This work has proposed the computational framework for mining causality from continuous-time dynamics models. To this end, we adapt the definition of neural Granger causality and propose a training algorithm that promotes sparse causal structure in the parameter space. We evaluate the effectiveness of our method with numerical experiments on two canonical benchmark problems: the Lorenz-96 and Mackey--Glass equations. We demonstrate our method's capabilities in accurate dynamics modeling and causality structure identification. We finally present an application of our proposed method to tsunami forecasting. The experimental results show that our method produces highly accurate tsunami forecasting (\textit{i.e.}, the predictions match the highest peaks of sea-surface elevation). Our method also identifies the causal structure of the time series obtained from the three gauges, which are physically-consistent and agree with the domain expert's interpretation.

\section{Acknowledgement}
Kookjin Lee (kookjin.lee@asu.edu) is the corresponding author and acknowledges the support from the U.S. National Science Foundation under grant CNS2210137.

{
    \balance
    \bibliographystyle{siam}
    \bibliography{ref}
}

\nobalance
\appendix

\section{Preliminaries on ANODEs or ANDDEs}
\label{appendix:prelim-anodes}

Here, we provide the background knowledge of augmented neural ordinary differential equations (ANODE) or neural delay differential equations (ANDDEs). ANODEs lift NODEs by adding arbitrary extra dimensions in the state variables leading to a system of ODEs, formally expressed as follows:
\begin{equation}
    \left[
    \begin{array}{c}
        \dot{{\LatentState}} \\
        \dot{\bm{s}}
    \end{array}
    \right]
    = \Velocity_{\NNParams} \left(  \left[
    \begin{array}{c}
        \LatentState \\
        \bm{s}
    \end{array}\right]\right).
\end{equation}
Although we have not numerically tested this in this study, augmenting the extra dimensions in NDDEs (ANDDEs) can be implemented by extending the NDDEs to take extra input variables as follows: 
\begin{equation}
    \left[
    \begin{array}{c}
        \dot{{\LatentState}} \\
        \dot{\bm{s}}
    \end{array}
    \right]
    = \Velocity_{\NNParams} \left(  \left[
    \begin{array}{c}
        \bar{\LatentState} \\
        \bm{s}
    \end{array}\right]\right)
\end{equation}
where the initial function of the augmented variables defined as $\bm{s}(t) = [s_1(t),\ldots,s_{n_s}(t)]\Transpose = \bm 0$ if $t\leq 0$. This 
is 
a straightforward extension of ANODEs in NDDEs. 

\section{Experimental Setup in Detail}
\label{appendix:exp-setup-detail}

\paragraph{Lorenz-96 system.}
%
In Sec.~\ref{subsec:lorenz96},
we set the forcing term as $F=10$, which causes a chaotic behavior.
The initial condition is set as $\pmb{x}(0) = [1,8,\ldots,8]$, perturbed by adding a noise which is sampled from uniform distribution $U(-1,1)$. The ground-truth trajectories are generated by solving the initial value problem with the time-step $\Delta t=0.01$ for the total simulation time $T=30.72$. For the time integrator, we use the Dormand--Prince method (dopri5)~\cite{dormand1980family} with the relative tolerance of $10^{-7}$ and the absolute tolerance of $10^{-9}$. 
We set the training hyperparameters as follows:
\begin{itemize}[itemsep=0.1em, topsep=0.1em]
    \item Learning rate: 0.01,
    \item Max epoch: 2000, 
    \item Batch size: 40, and
    \item Batched subsequence length: 100.
\end{itemize}

\paragraph{Mackey--Glass system.}
In Sec.~\ref{subsec:mg}, 
we set $a=0.2$, $b=0.1$, $c=10$, and $\tau=5$ (seconds). We solve the initial value problem, given the initial function $\phi(t)=.5$, using \textsc{ddeint}\footnote{\url{https://pypi.org/project/ddeint}} with the time-step $\Delta t = 0.01$ for the total simulation time $T=153.6$ (seconds). 
We set the training hyperparameters as follows:
\begin{itemize}[itemsep=0.1em, topsep=0.1em]
    \item Learning rate: 0.01,
    \item Max epoch: 2000, 
    \item Batch size: 40,
    \item Batched subsequence length: 100, and
    \item ODE integrator: Runge--Kutta~\cite{runge1895numerische} of order 4.
\end{itemize}

\paragraph{Tsunami forecasting.}
We set the training hyperparameters as follows:
\begin{itemize}[itemsep=0.1em, topsep=0.1em]
    \item Learning rate: 0.001,
    \item Max epoch: 1000,
    \item Batch size: 40,
    \item Batched subsequence length: 10, and
    \item ODE integrator: Runge--Kutta of order 4.
\end{itemize}

\end{document}